\pdfoutput=1

\documentclass{article}
\usepackage{jfrExamplee}
\usepackage{ifpdf}
\usepackage{graphicx}
\usepackage{url}
\usepackage[lofdepth,lotdepth]{subfig} 
\usepackage[export]{adjustbox} 
\usepackage{apalike}
\usepackage{setspace}
\usepackage{algorithm}
\usepackage[noend]{algpseudocode}
\usepackage{amsmath} 
\usepackage{amssymb}  
\usepackage{amsfonts}
\usepackage[title]{appendix}

\makeatletter
\let\OldStatex\Statex
\renewcommand{\Statex}[1][3]{%
  \setlength\@tempdima{\algorithmicindent}%
  \OldStatex\hskip\dimexpr#1\@tempdima\relax}
\makeatother

\title{Adaptive Path Planning for Depth Constrained Bathymetric Mapping with an Autonomous Surface Vessel}

\author{
Troy Wilson \\
Australian Centre for Field Robotics\\
The University of Sydney\\
Sydney, New South Wales, Australia \\
\texttt{t.wilson@acfr.usyd.edu.au} \\
\And
Stefan B. Williams \\
Australian Centre for Field Robotics\\
The University of Sydney\\
Sydney, New South Wales, Australia \\
}

\begin{document}
\maketitle
\begin{abstract}
This paper describes the design, implementation and testing of a suite of algorithms to enable depth constrained autonomous bathymetric (underwater topography) mapping by an Autonomous Surface Vessel (ASV). Given a target depth and a bounding polygon, the ASV will find and follow the intersection of the bounding polygon and the depth contour as modeled online with a Gaussian Process (GP). This intersection, once mapped, will then be used as a boundary within which a path will be planned for coverage to build a map of the Bathymetry.  Methods for sequential updates to GP's are described allowing online fitting, prediction and hyper-parameter optimisation on a small embedded PC. New algorithms are introduced for the partitioning of convex polygons to allow efficient path planning for coverage. These algorithms are tested both in simulation and in the field with a small twin hull differential thrust vessel built for the task.
\end{abstract}

\section{Introduction}

Navigational maps are important for the safe passage of recreational and commercial boating traffic. Traditionally these are created with sonar data collected from surveying vessels. The high cost of conducting these surveys impacts the frequency of re-surveying  \cite{Senet2008}. An Autonomous Surface Vessel (ASV) able to conduct these surveys autonomously could significantly reduce this cost and thus enable more frequent surveying to occur for a fixed budget due to lower equipment and personnel costs. An additional benefit arises from the fact that smaller draught and reduced thrust of the ASV will allow surveying in shallower waters and with less disturbance in sensitive environments. 

This paper describes a system of algorithms tested in simulation and implemented in the field which demonstrate the use of an autonomous system for bathymetric mapping. A small twin hull, differential thrust ASV with a single beam sonar for depth sensing and GPS and IMU for localisation is modeled and built. The vessel is shown to autonomously build a model of the bathymetry with a Gaussian Process (GP), where the model is continuously fit online as data arrives and the parameters of the model are periodically re-estimated. A minimum depth target combined with a bounding polygon is used to both avoid static obstacles and define the area to be explored. A control algorithm plans a path to follow the intersection of the depth contour and bounding polygon. Once this boundary is mapped a lawnmower path is then planned within it for coverage. A new algorithm based on the Boustrophedon Cellular Decompostion (BCD) is developed to achieve this. To allow this system to run in real time on a small embedded system in the field, incremental algorithms have been adopted for the covariance matrix updates and the calculation of their Cholesky decompositions required for the GP model. A multi-threaded implementation has been used to share the computational load between the 2 physical (4 virtual) cores on the ASV. The outcomes of this work are validated both in simulation and in a deployment of the ASV in an estuarine environment.

The remainder of this paper is organised as follows. Section 2 presents related work, reviewing the current state of the art in autonomous route planning. Section 3 provides a brief summary of Gaussian Processes. Section 4 details the algorithmic suite developed to enable the autonomous bathymetric surveying. Section 5 and 6 then test these algorithms in simulation and the field. Section 7 summarises the work with conclusions and avenues for future research.

\section{Related Work}

Current robotic surveying work often involves pre-planned survey paths which require prior information on the area to be surveyed and cannot react to information as it is received \cite{Clark2013a,Grasmueck2006,JohnsonRobertson2010}. Creating the optimal back and forth path for coverage of an area whilst staying within the workspace, which is referred to as a lawnmower path in the robotics literature \cite{Galceran2013} or an axis parallel solution to the milling problem in computational geometry, has been shown to be related to the Travelling Salesman Problem (TSP) and thus NP hard in general \cite{Arkin2000}. By partitioning the complex workspace into a number of simpler shapes, which can easily be solved, and then joining these spaces together, it is possible to produce feasible paths for coverage in polynomial time. The joining together of these cells themselves optimally is also a TSP problem, and thus approximations must be used here as well for polynomial time solutions.

A polygonal workspace can be spilt into its elemental trapezoids, known as the Trapezoidal decomposition, \cite{latombe2012robot}. These trapeziods are convex, and thus lawnmower paths in any direction can cover the space. This method whilst simple to implement, can result in an excessive number of elemental cells, which can lead to a large number of inefficient transit paths to join these together. Some of these cells could be merged back together to create larger elemental convex polygons and then joined as shown in \cite{Oksanen2009}. The authors of that work also implement a heuristic for optimising the orientation of the tracks within each cell. Alternatively, a sweep direction can be chosen through the polygon, which is then split into elemental polygons which are monotone to this sweep direction. A polygon is monotone to a sweep direction if lines orthogonal to this sweep direction cross the polygon at most twice. This guarantees by construction that these elemental polygons can be covered by tracks orthogonal to the sweep direction, however this also means that all cells must be covered by tracks in the same orientation. This method was introduced as the Boustrophedon Cellular Decompostion (BCD), \cite{Choset1998}. It results in a smaller number of elemental cells than even the merged version of the Trapezoidal Decomposition, thus reducing the number of transit paths between them. This is not guaranteed to reduce the total transit length, especially if using heuristics to solve the transit paths, however reducing the number of cells is especially important for online implementations given the optimal solution is NP in relation to the number of cells. A graph of the elemental cells and their neighbours is created during the decomposition, and a simple search through the graph to the next anti-clockwise cell is conducted to join the cells together. This approach has been extended to non-polygonal workspaces and non-linear sweep lines as the Morse Decomposition \cite{Acar2002}. The BCD has also been expanded to work online in unstructured environments. Whilst some simple rules are used to allow for rejection of bad data, all other data point are assumed to be perfectly sensed \cite{Acar2001}.

The advantage of adaptively reacting to sensed information has been demonstrated \cite{Ferri2008}. The authors used a chemical sensing threshold to trigger a detour from a coarse survey path into a fine spiral for increasing coverage in areas likely to contain hydro-thermal vents. This was rather brittle however, as the threshold was manually set a priori. The number of trigger events was limited, with no analysis done on where they would be most usefully deployed. This method leads to the possibility of using up available spirals too early in the mission and missing valuable information later.

Using the sensor data as it is collected to build a model of the parameter of interest to be studied and then planning within this model offers the potential for surveying in unknown environments under uncertainty and for optimising paths taken to produce models based on levels of certainty and resource use. GP's offer a useful framework to deal with estimation under uncertainty and planning and can be conducted on both mean and variance information. There is a growing literature on planning within GP's, for the problem of sensor placement, which ignores travel times \cite{Krause2008}. More recently research has focused on adaptively planning for a mobile sensor platform which reacts to sensor data as it is received \cite{Souza2014,Marchant2012,Yang2013,Bender2010,Lawrance2011,Nguyen2013,Marchant2014,Hollinger2012}, though few have implemented this online in the field.

There have been a number of published studies on the design, development and testing of small autonomous surface vessels for robotics research in recent years. Twin hull vessels have the advantage of being relatively stable in roll and have been implemented by a number of authors 
\cite{hitz2012,Matos2008,Soares2013,Tokekar2010}. All but the last of which used differential thrust for increased manoeuvrability, allowing rotation on the spot. Some examples of larger autonomous surfaces vessels are the full scale catamaran used for methane sensing on an inland dam \cite{Dunbabin2009} , the MIT  AutoCat and Kayaks \cite{Manley2000}, Swordfish \cite{Ferreira2007}, Delfim \cite{Pascoal2000} and the Springer USV \cite{Naeem2012}. There has also been some commercial development in  ASV's such as the Wave Glider by Liquid Robotics \cite{hine2009wave} and the  Saildrone \cite{Saildrone}.

These vehicles are generally underactuated and nonholonomic. This reduced number of degrees of freedom in their action space requires trade-offs to be made in the control algorithms. The environments they are deployed in also generally have external forces in the form of winds and currents acting on the vessel. Station keeping of a vessel in the presence of external forces has been shown to be possible with thrust control proportional to distance to target and yaw control using a proportional and integral controller \cite{Matos2008} and a full PID controller has been demonstrated for speed and line following \cite{hitz2012}. 

In this work we show how building a model of the bathymetry and simultaneously planning within this model allows exploration of the intersection of a depth contour and a bounding polygon in an unknown environment under uncertainty. We present the Discrete Monotone Polygonal Partitioning algorithm to produce elemental cells that can be exactly covered by the desired lawnmower path width thus resulting in more efficient paths than the BCD, and then fit a path through these cells for  coverage of this area to build a model of the bathymetry. We implement methods for efficient updating of the GP to allow online prediction of the bathymetric profile and estimate and update the GP hyper-parameters as we collect data. This is tested in both in simulation and in the field with a small low cost ASV.

\section{Gaussian Processes}
\label{GP}
Guassian Process regression is a powerful method for estimating a spatial process. It is a covariance based procedure where inference is strong close to observed data points, but as the distance from observed data increases the confidence of the prediction drops \cite{Williams2006}. The form of the covariance function $k(',')$, must be specified and a common form is the Squared Exponential Covariance function:
\begin{equation} \label{eq:SqExp}
k(\mathbf{x_i},\mathbf{x_j}) = \sigma_f^2\text{exp} \left(-\frac{|\mathbf{x_i}-\mathbf{x_j}|^2}{2l^2}\right)
\end{equation}
\noindent where $x_i, x_j$ are the 2 observations, $\sigma_f^2$ is the process noise and $l$ the characteristic length scale. Under a Gaussian Process it is assumed that the joint distribution of training points $\mathbf{x}$ with realisations $\mathbf{y}$ and test points $\mathbf{x_*}$ with realisations $\mathbf{y_*}$ is jointly normal: 
\begin{equation} \label{eq:GPDist}
\begin{bmatrix}
\mathbf{y} \\
\mathbf{y*} \\
\end{bmatrix}
\sim \mathcal{N}
\left(
\mathbf{0},
\begin{bmatrix}
K + \sigma_n^2I & K_*^T\\
K_* & K_{**}\\
\end{bmatrix}
\right)
\end{equation}
\noindent where $\sigma_n^2$ is the observation noise and $I$ is the identity matrix. The expectation and variance of the test points are given by:
\begin{equation}\label{eq:yHat}
\hat{\mathbf{y}}_* = K_*K_y^{-1}\mathbf{y}\\
\end{equation}
\begin{equation}\label{eq:sigHat}
\hat{\mathbf{\Sigma}}_* = K_{**} - K_*K_y^{-1}K_*^T
\end{equation}

\noindent where:
\[
K = K(X,X) \quad K_* = K(X_*,X) \quad K_{**} = K(X_*,X_*) \quad K_y = K+\sigma_n^2I
\]
Here we are using the convention of lower case letters for scalars or vectors and upper case for matrices. There are 3 hyper-parameters ($\theta$) which must be set, namely $\sigma_n^2$, $\sigma_f^2$ and $l$. These hyper-parameters can be set from prior data or they can be estimated online. A fully Bayesian treatment would marginalise out the hyper-parameters, however for computational speed Maximum-Likelihood (ML) estimation is often used. To conduct ML we need a function for the likelihood of the  data given the hyper-parameters. For computational stability reasons the log transformation of this  number, the Log Marginal Likelihood (LML) is often used and is defined as:

\begin{equation} \label{eq:LML}
\text{log}p(\mathbf{y}|X)= -\frac{1}{2}\mathbf{y}^TK_y^{-1}\mathbf{y} -\frac{1}{2}\text{log}|K_y| - \frac{\text{n}}{2}\text{log}2\pi
\end{equation}

Cholesky factorisation is generally used instead of the direct matrix inversion required in Equations \ref{eq:yHat} and \ref{eq:sigHat} for both numerical stability and speed. Even with this however, the calculation of $K$ when fitting the GP is $O(n^3)$. Once we have $K$, the partial derivatives of the LML w.r.t $\theta$ are relatively inexpensive to calculate. This allows the use of fast numerical optimisation techniques such as the Limited memory Broyden Fletcher Goldfarb Shanno algorithm with bound constraints (L-BFGS-B),  which require these gradients to conduct the ML estimation of the hyper-parameters \cite{zhu1997algorithm}.

\begin{equation} \label{eq:dLMLdTheta}
\frac{\partial}{\partial\theta_i}\text{log}p(\mathbf{y}|X) = \frac{1}{2}\text{tr}\left((\alpha\alpha^T-K_y^{-1})\frac{\partial K_y}{\partial\theta_i}\right)  \quad\quad  \alpha = K^{-1}\mathbf{y}
\end{equation}

\begin{equation} \label{eq:dKdThetai}
\frac{\partial K_y}{\partial \sigma_f^2} = \text{exp}\left(-\frac{\|r\|^2}{2l^2}\right)
\quad
\frac{\partial K_y}{\partial \sigma_n^2} = I
\quad
\frac{\partial K_y}{\partial l} = \frac{\|r\|^2}{l^3}\sigma_f^2\text{exp}\left(-\frac{\|r\|^2}{2l^2}\right)
\end{equation}

\begin{figure}[t]
\centering
\includegraphics[scale = 0.5, natwidth=615 , natheight=615]{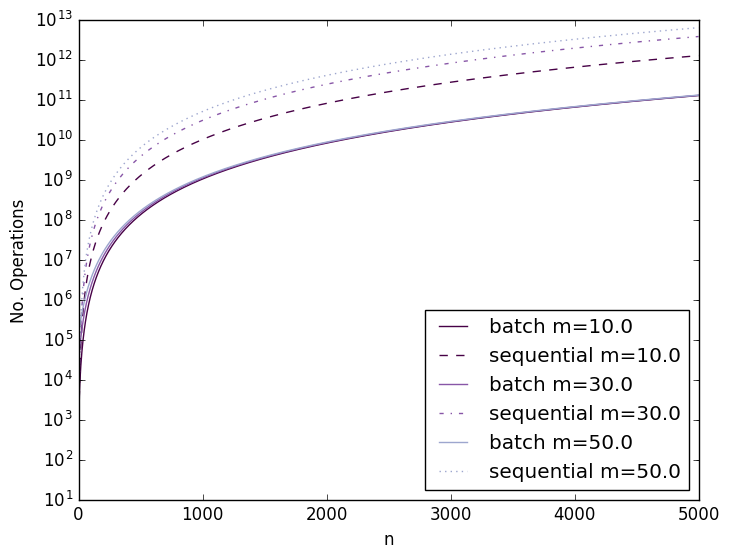}
\caption{Operations Count for Batch vs. Sequential GP Prediction}
\label{fig:Ops}
\end{figure}

When we wish to make predictions from the GP with $n$ training points $\mathbf{x}$ and $m$ test points $\mathbf{x_*}$ we would calculate this as a batch and thus be dominated by the calculation of $K, K_*$ and $K_{**}$ which would be $O((n+m)^3)$. For computational efficiency, if multiple predictions are to be made on the same set of test data, care should be taken to design algorithms such that all predictions are made in one batch. For instance predicting in a batch would be $O((n+m)^3)$ rather than $O(m*(n+1)^3)$ if calculated sequentially, which in the case of n=500 and m=50 would be 38x faster. This can be seen in Figure \ref{fig:Ops} where the number of operations for the sequential calculations is more than an order of magnitude slower as n grows.

\section{Autonomy Suite of Algorithms}
The aim of our study is to provide a suite of algorithms to allow an ASV to operate in an unstructured environment, with minimal prior information, and to autonomously explore the area and return a map of the bathymetry. 

To achieve this there are three main components to the algorithm. Firstly we have the GP which is updated with data as it is collected by the ASV. For this to run online on a small embedded CPU care must be taken both in how the GP is queried, as discussed in Section \ref{GP}, and also in how the GP is updated. We are also estimating the hyper-parameters of the GP from this data online with ML estimation of the LML. Secondly, we have developed an algorithm to follow the intersection of a bounding polygon and the depth contour as predicted by the GP. Thirdly, once this has concluded we propose an efficient Discrete Monotone Polygonal Partitioning (DMPP) of the resultant intersection polygon. We then sequentially plan a transit to the closest cell followed by a lawnmower path within it, and repeat until all cells are covered. The ASV then follows this path. 

\subsection{Online Gaussian Process Updates}
As discussed in Section \ref{GP}, for $n$ training points and $m$ test points, fitting of a GP is $O(n^3)$ and prediction from the GP is $O((n+m)^3)$. This is due to the calculation of the Cholesky factorisation of the covariance matrix. When we are adding data, or predicting $m$ test points, instead of recalculating the entire Cholesky factorisation, we can simply calculate the extra columns and rows related to the new data/test points, and add these to the matrix we have already calculated. For $m$ new training/test points this reduces the update step from $O((n+m)^3)$ to the larger of $O(n^2)$ or $O(m^3)$, due to Equations \ref{eq:S12} and \ref{eq:S22} respectively. This can be done exactly for the case of additions of data. Outside of robotics GP's are generally used on a batch of data once it has been collected. Optimisations in software implementations of GP's focus on sparsifying or reducing the size of the covariance matrix to reduce computation time on one batch of data. The authors are not aware of any GP packages which implement sequential data updates and as such we will briefly detail this procedure for updating the Cholesky matrix. This is a simplified version of the method detailed in \cite{Osborne2008} which provides algorithms for both updates and downdates (which are not exact) anywhere in the Cholesky matrix.

We have the positive semi-definite (p.s.d) covariance matrix $K_{11}$ and its upper triangular Cholesky matrix $L_{11}$. If we are adding new data points to the end of the covariance matrix we then have 
$
\begin{bmatrix}
K_{11} & K_{12} \\
K_{12}^T & K_{22} \\
\end{bmatrix}
$
. The new elements $K_{12}$, its transpose $K_{12}^T$ and $K_{22}$ are calculated from the new data or test points using Equation \ref{eq:SqExp}. We want to calculate the resulting Cholesky Matrix
$
\begin{bmatrix}
S_{11} & S_{12} \\
0 & S_{22} \\
\end{bmatrix}
$
. For a triangular $\mathbf{A}$ we can use backwards substitution to solve $\mathbf{Ax = b}$, defined as $\mathbf{x = A\backslash b}$. We thus find the following solutions for the elements of $S$:

\begin{equation} 
S_{11} = L_{11}
\end{equation}
\begin{equation} \label{eq:S12}
S_{12} = L_{11}^T\backslash K_{12}
\end{equation}
\begin{equation} \label{eq:S22}
S_{22} = \text{chol}(K_{22} - S_{12}^T S_{12}) \quad \text{or for } m=1  \quad S_{22} = \sqrt{K_{22}}
\end{equation}

where chol() signifies the Cholesky decomposition. The GP is run on its own thread and its covariance and Cholesky matrices are updated as sonar data arrives using the sequential method described above. When predictions are required from the GP, the current covariance matrix of training points is taken, the new covariance for the test points with themselves, $K_{22}$, and with the training points, $K_{12}$, are calculated and then the sequential method is used to update the  Cholesky matrix to solve the GP for $\mathbf{y^*}$.

The GP also requires determination of its hyper-parameters. As stated in Section \ref{GP}, this can be done using ML estimation of the LML. We use the L-BFGS-B algorithm with the analytical gradients calculated in Equations \ref{eq:dKdThetai} and \ref{eq:dLMLdTheta}. This is run after an initialisation period to collect some data and then at regular intervals. It is run on a separate thread, and when it returns new hyper-parameters these are then used by the GP. It is important to note that these hyper-parameters are used to define the covariance matrix, and thus when they change, both the covariance matrix and the Cholesky matrix need to be fully recalculated, before performing any new sequential data updates or predictions.

\subsection{Find and Follow the Intersection of a Depth Contour and a Boundary}

The adaptive autonomy of the ASV is provided by the algorithm for finding the desired depth contour, defined by a target depth $z_t$ and following the intersection of this with the bounding polygon, which is detailed in Algorithm \ref{alg:FFCB}. The depth contour is set at the intersection of a safe operating depth and a minimum depth of interest for the study. For a point sensor only sensing directly downwards we cannot avoid floating objects such as bouys, or very steep gradients such as vertical seafloor rises. The vessel should be able to both operate safely due to these bounds and obtain the sonar data required to create a bathymetric map of the area. The bounding polygon will keep the ASV both in an area of interest and away from obstacles unable to be predicted through modelling of the bathymetry. Inspired by Bug type algorithms such as DistBug \cite{Kamon1997}, the algorithm follows the desired depth contour until it hits a boundary. Upon hitting the boundary it then follows this boundary until it finds the boundary taking it shallower than the target depth, at which point it leaves the boundary and again begins following the depth contour. This is continued until a circuit has been completed. A key difference in our algorithm to the bug algorithms is that we are not aiming for a single goal but always searching for  a desired depth at a distance $r$, the search radius from our current position. This search is detailed in Algorithm \ref{alg:RoseSolve}. In addition, the surface upon which we are searching, the GP model of the bathymetry, is changing as we obtain data and both refit the model and re-estimate the parameters of the model. This model changes faster in the initial stages when we have small amounts of localised data. This can lead to the initial path turning back on itself. To counteract this noise, and prevent early determination of boundary closure, we set a parameter on the boundary completion test for loop closure to ignore the most recent $loopBu\hspace{-0.1em}f\hspace{-0.2em}f\hspace{-0.1em}er$ points. The value of $loopBu\hspace{-0.1em}f\hspace{-0.2em}f\hspace{-0.1em}er$ should be set relative to the expected length of the boundary. 

\begin{figure}[t]
\centering
\includegraphics[scale = 0.7, natwidth=896 , natheight=558, trim=0cm 0cm 0cm 0cm]{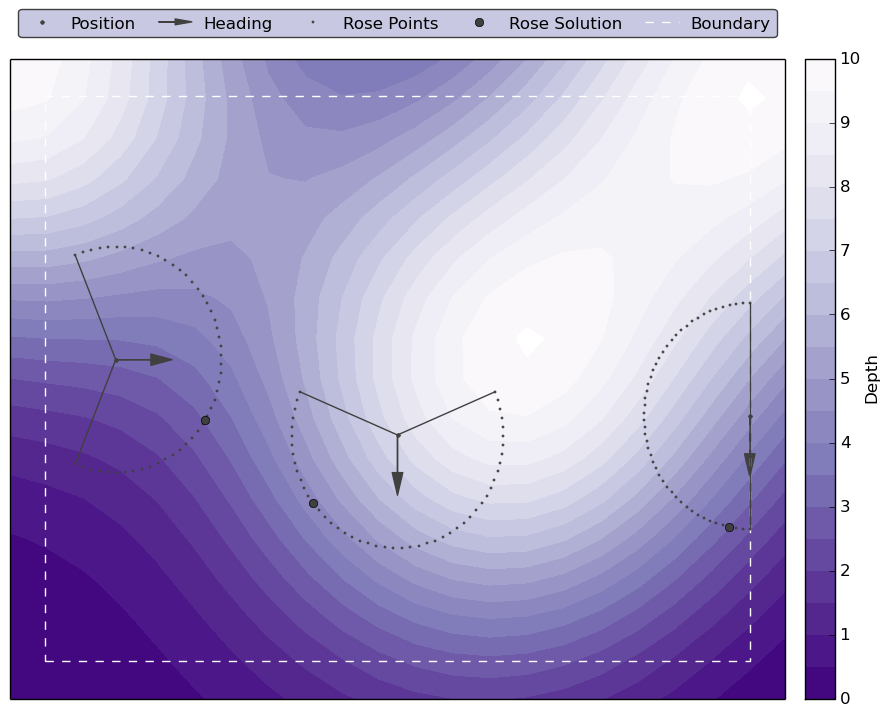}
\caption{Example of RoseSolve Algorithm to Solve for Desired Heading Based on Target Depth of 3}
\label{fig:roseSolve}
\end{figure}

\begin{algorithm}
\caption{Find and Follow Contour within Bounding Polygon}
\label{alg:FFCB}
\begin{algorithmic}[1]
\Procedure{FFCB}{$poly, r, z_t, loopBu\hspace{-0.1em}f\hspace{-0.2em}f\hspace{-0.1em}er, \psi_{adj}$}
\State $mode$ = Contour
\State $b, f\hspace{-0.1em}oundCountour = \emptyset, \emptyset$
\State $bCont = []$
\Repeat
	\State $GP$ = getGP()
	\State $x, y, \psi$ = getPose()
	\If{$mode$ = Contour}\ \Comment{Contour Mode}						
		\State $\psi_s = \psi-\psi_{adj}$
		\State $\psi_e = \psi+\psi_{adj}$
		\State $x_w, y_w$ = roseSolve($GP,z_t,\psi,x,y,r,\psi_s,\psi_e$)
		\If{pointInPolygon($x_w, y_w$) = True}
			\State $\psi_d$ = atan2($x_w-x, y_w-y$)
		\Else\ \Comment{Crossing boundary}
			\State $mode$ = Boundary \ \Comment{Switch to Boundary Mode}			
			\State $edge$ = getEdge($[x,y],[x_w,y_x]$)
			\If{$b = \emptyset$}
				\State $b$ = setDirection()
			\EndIf
			\State $x_w, y_w$ = getVertex($b, edge$)
			\State $\psi_d$ = atan2($x_w-x, y_w-y$)
		\EndIf
	\Else\ \Comment{Boundary Mode}	
		\State $x_w, y_w$ = getVertex($b, edge$)
		\If{$\|[x_w, y_w],[x,y]\| < r$}
			\State $edge = $getNextEdge$(b, edge$)
			\State $x_w, y_w$ = getVertex($b, edge$)
		\EndIf
		\If{Depth($x_w, y_w$) $> z_t$}
			\State $\psi_d$ = atan2($x_w-x, y_w-y$)
		\Else \ \Comment{Leave boundary}
			\State $mode$ = Contour \ \Comment{Switch to Contour Mode}
			\State $\psi_s, \psi_e$ = arcPoly($x,y,poly$)	
			\State $x_w, y_w$ = roseSolve($GP,z_t,\psi,x,y,r,\psi_s,\psi_e$)
			\State $\psi_d$ = atan2($x_w-x, y_w-y$)
		\EndIf	
	\EndIf
	\State $z$ = depth($\psi, r, x, y$) 
	\If {$f\hspace{-0.1em}oundCountour = \emptyset$}
		\If {($mode$ = Boundary) or 
		\Statex[3]$\lvert z-z_t\rvert <\epsilon$)}
			\State $f\hspace{-0.1em}oundContour$ = True
		\EndIf
	\EndIf
	\If {$f\hspace{-0.1em}oundCountour$ = True}
		\State $bCont$.append($x, y$)
	\EndIf	
	\State bendMotorCommand($\psi_d$)	
\Until{boundaryComplete($bCont, x,y,loopBu\hspace{-0.1em}f\hspace{-0.2em}f\hspace{-0.1em}er $)}\\
\Return{$bCont$}
\EndProcedure
\end{algorithmic}
\end{algorithm}

\begin{algorithm}
\caption{Solve for heading to target depth $t_z$}
\label{alg:RoseSolve}
\begin{algorithmic}[1]
\Procedure{roseSolve}{$GP, z_t, \psi, x ,y , r, \psi_s, \psi_e$}
\State $Splits = 50$
\State $\Delta\psi = $arcLen$(\psi_s, \psi_e)/Splits$
\State $\psi_*[0] = \psi_s$
\For{$i$ in $1$ to $Splits$}
	\State $\psi_*[i] = \psi_*[i-1] + \Delta\psi$
\EndFor
\State $ x_* [], y_* [] = $cartesian($x, y, \psi_* [],r$)
\State $z_*[] $ = depthBatch($GP, x_*, y_*$)
\For{$i$ in $Splits$}
	\State $j = $Mod$((i+1), Splits)$
	\State $\psi [i], z [i] $ = bestHeading($ \psi_* [i], \psi_* [j], z_* [i], z_* [j], z_t$) 
	\State $\psi_{\epsilon} [i] = $abs$(\psi [i] - \psi )$
	\State $z_{\epsilon} [i] = $abs$(z [i] - z_t )$
\EndFor
\State $\psi []$ = sort $\psi []$ by $z_{\epsilon}, \psi_{\epsilon}$
\State $x_w, y_w$ = cartesian($x, y, \psi [0],r$)\\
\Return{$x_w, y_w$}
\EndProcedure
\end{algorithmic}
\end{algorithm}

The RoseSolve() algorithm searches for the heading leading to the target depth on the circumference of an arc on the compass rose of radius $r$ around the current position, as shown in Figure \ref{fig:roseSolve}. Whilst in contour following mode, this arc is centred around the current heading and bounded by the start angle, $\psi_s$ and end angle,$\psi_e$. This range, $\pm \psi_{adj}$ can be chosen to reduce the search space so we are not wasting computation searching where we have just come from. When exiting boundary following mode, $\psi_s$ and $\psi_e$ for the search are set based on the boundary of the polygon, such that the arc is inside the polygon. This arc is then evenly split into a number of points, which are converted into Cartesian co-ordinates and the predicted depth for these positions is returned in a batch from the GP. Linear interpolation is conducted between each pair of points for the heading which returns the depth closest to the target depth through the function bestHeading(). We then return the best of these solutions with ties broken based on the distance to our current heading. This behaviour can be seen in Figure \ref{fig:roseSolve}, where the algorithm is solving for a desired depth of 3. For the position on the left and near the middle, an arc centred around the current heading is searched, whilst in the position on the right boundary it can be seen we are searching on the arc within the boundary. The depth for all the points on these arcs are queried from the GP, and then the best segment linearly interpolated to get the solution, indicated by the solid grey dot. For the positions on the left and right, the algorithm finds the heading which will lead to the desired depth, whereas the position in the middle returns the heading which is closest to the desired depth.

Originally we implemented this as a recursive bi-section search on the arc, however the computational cost of repeatedly querying the GP for 1 prediction point method led us to implement a batch procedure. The number of splits is a design parameter. With 50 splits, we have a prediction point no more than $1/40\pi$ radians apart, which with a search radius $r = 5$m equates to test point spacing of approximately 0.4m. The velocity and sampling frequency on our ASV resulted in spacing of the sampling of points of around 1m along the path of the vehicle. Thus the combination of this test point spacing with linear interpolation, the smooth surface provided by the GP on this scale given the sampling scale, and the  frequency on the control loop speed was found to be a reasonable compromise for computational load and precision.

A key tunable parameter in this algorithm is the search radius. A number of factors come into play in the choice of the search radius. The smaller this radius is, the smaller the area in the GP around the current position of the vessel is searched for the desired heading. A larger radius will result in a smoother path, however, this will also result in a larger tracking error between the desired depth found on this radius, and the depth sensed directly below the vessel. The expected rate of change of the bathymetry should be taken into account such that the search radius is set to allow the vessel to follow these changes. The speed of the vessel and the sampling rate of the sonar should also be taken into account.

\subsection{Discrete Monotone Polygonal Partitioning and Path Generation}
Upon completion of the intersection of the depth contour and the bounding polygon we have a new intersection polygon around which we have already sensed and which we now need to plan within for coverage. For the purpose of this study the track width is a design parameter. Similar to the BCD we have implemented a method to create elemental polygons which are Monotone to a given sweep direction. There are some differences however which we have implemented to produce a more optimal path given the desired path spacing.

\begin{figure}[h]
\centering
\includegraphics[scale = 0.6, natwidth=1400 , natheight=600, trim=4.5cm 0cm 0cm 0cm]{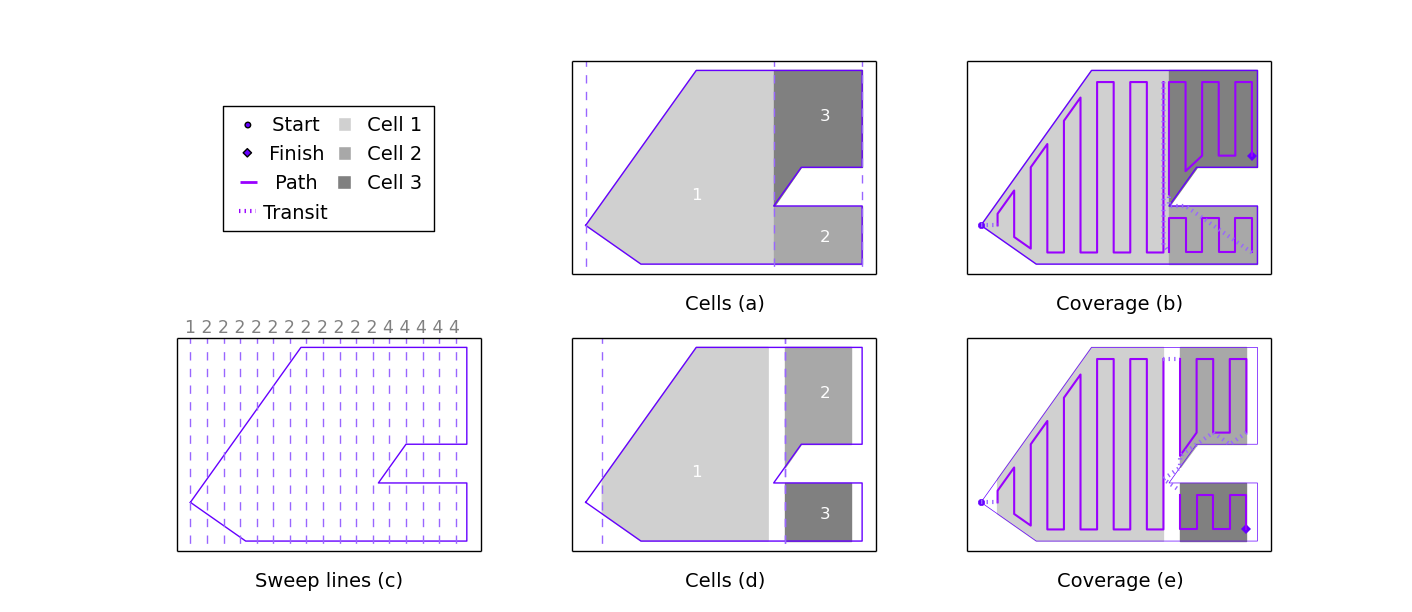}
\caption{Comparison of BCD (a,b) with DMPP (c,d,e) }
\label{fig:boustComp}
\end{figure}

Under BCD, a sweep direction is chosen. A line orthogonal to this sweep direction is traced through the polygon. At any point where the number of crossings of this line with the polygon changes we have a connectivity event. These events are used to create the elemental polygons which are monotone to this sweep direction. This can be seen in Figure \ref{fig:boustComp}(a), where there are 3 changes in the connectivity count which is used to create the three elemental polygons (note this is using a horizontal sweep direction left to right). The union of these three polygons returns the original polygon. These polygons are then joined in an anti-clockwise order as shown by Figure \ref{fig:boustComp}(a), with lawnmower paths then drawn within them as shown by Figure \ref{fig:boustComp}(b). 

The first issue with this partitioning is that the width of these polygons is not necessarily a multiple of the track width. The result of this can be seen in Figure \ref{fig:boustComp}(b), where the last up track in cell 1 is close to the cell boundary. The first vertical tracks in cells 2 and 3 are then closer to this track than our desired spacing which results in longer total path length and irregular coverage. For our application of a point sensor this results in some sensed points being closer than required. In the application from \cite{Choset1998} of perfect sensing of a fixed width scanner, this would result in a significant amount of scan overlap on the cell edges, again a waste of resources. We have developed the Discrete Monotone Polygonal Partitioning(DMPP) method to explicitly deal with this, which will be discussed after detailing the other shortcomings of the BCD method.

The second issue with the BCD method is that the transit paths joining the cells are not optimised. The order of joining is simply conducted through an anti-clockwise search of the neighbours of the current cell for the first cell that has not yet been covered, repeated until all cells are covered. It is not until all the transit paths are computed that the lawnmower path within the cell is computed. This is clearly sub-optimal. As can be seen in Figure \ref{fig:boustComp}(b), lawnmower coverage of cell 1 ends in the top right hand corner. Clearly the best cell to go to would be the upper right cell, however the BCD method is not aware of this and instructs a transit to the bottom right cell. Which corner of the cell to join to should also be dependent on which corner of the prior cell we exited from, however the BCD method does not actually detail how the lawnmower paths within the cell should be constructed, which in any case are only fit after the order of cell joining is determined. We deal with this second issue by calculating the lawnmower path within the closest cell, then solving an A* search from the exit point of this cell to the nearest corner of an unexplored cell, and repeating until all cells are covered. 

Finally the BCD is designed to handle polygonal objects within its boundary. The boundary is set up as a simple rectangle. Whilst there is nothing to stop their general approach being applied to more complex boundaries, they do not mention this case and how it would be approached. Our algorithm explicitly deals with this case.

Algorithm \ref{alg:DMPP} details the DMPP. The input parameters for this algorithms are the boundary polygon $poly$, the desired track width $\delta$ and the sweep direction $\psi_{sd}$. Without loss of generality, by constraining the sweep direction in the range $-\pi/2$ to $\pi/2$, we can sweep from left to right from the bottom left corner defined by $x_{min}$ and $y_{min}$. Lines 6-11 of Algorithm \ref{alg:DMPP} detail this sweeping process of counting the number of crossings, and their position of a line orthogonal to the sweep direction with the polygon. The list of positions is returned sorted from the bottom of the sweep line. An example of this can also be seen in Figure \ref{fig:boustComp}(c) where the sweep lines are shown, and their corresponding crossing count shown on the top of the chart. 

Once we have this list of crossings, we then loop though this list looking for changes in the number of crossings from 1 sweep line to the next, indicating a connectivity change event. When this happens we close our open cells in order from bottom to top, before we then open new cells, again ordered from bottom to top. The co-ordinates of the individual cell corners created are ordered clockwise from the bottom left corner. These 4 points define the first and last tracklines in the cell. We then need to follow the boundary between point 2 to 3 and point 3 to 4 to fully define the elemental cell, through the primitive procedure traceBoundary(). This completes the creation of our Discrete Monotone Polygonal Partitioning.

\begin{algorithm}
\caption{Discrete Monotone Polygonal Partitioning (DMPP)}
\label{alg:DMPP}
\begin{algorithmic}[1]
\Procedure{DMPP}{$poly, \delta, \psi_{sd}$}
\State $x_{min}, x_{max}, y_{min}, y_{max} =$ Extremes$(poly)$
\State $\psi_{sl} = \psi_{sd} + \pi / 2$
\State $x, y = x_{min}, y_{min}$ \Comment{start sweep from bottom left corner}
\State $sweepCrossings, sweepCount = \emptyset, \emptyset$
\While{$x < x_{max}$  or  $y < y_{max} $} 
	\State $crossings = $RayXPoly$(x,y,\psi_{sl})$	\Comment{return a sorted array of the position of each crossing}
	\State $sweepCrossings.$append$(crossings)$ 
	\State $sweepCount.$append(len($crossings$)) 
	\State $x = x + $cos$(\psi_{sd})\delta$
	\State $y = y + $sin$(\psi_{sd})\delta$
\EndWhile

\State $openCellCnrs, cellCnrs = \emptyset, \emptyset$
\State $prevCount, cellCount = 0, 0$
\For{$i$ in len($crossings$)}
	\If{$(sweepCount[i] - prevCount)$ $!= 0$} \Comment{connectivity change event}
		\State $cc = 0$
		\While{len($openCellCnrs$) $>0$} 	\Comment{close open cells}
			\State $oc = openCellCnrs.$pop 
			\State $cellCnrs[oc][3] = crossingAr[i-1][cc]$		
			\State $cellCnrs[oc][2] = crossingAr[i-1][cc+1]$
			\State $cc = cc + 2$
		\EndWhile	
		\For{$j$ in int($crossingCount[i]/2$)} \Comment{open new cells}
			\State $openCellCnrs$.append($cellCount$)
			\State $cellCnrs[cellCount][0] = crossingAr[i][j*2]$		
			\State $cellCnrs[cellCount][1] = crossingAr[i][j*2+1]$		
			\State $cellCount = cellCount +1$
		\EndFor				
	\EndIf
	\State $prevCount = sweepCount[i]$
\EndFor
\For{$i$ in len($cellCnrs$)}
	\State $cells[i] = [cellCnrs[i][0:2]]$
	\State $cells[i].$append(traceBoundary($cellCnrs[i][1], cellCnrs[i][2], poly$))
	\State $cells[i].$append($cellCnrs[i][2:]$)	
	\State $cells[i].$append(traceBoundary($cellCnrs[i][3], cellCnrs[i][0], poly$))
\EndFor
\Return{$cells, cellCnrs$}
\EndProcedure
\end{algorithmic}
\end{algorithm}

Now that we have our DMPP, we can proceed to generate a path for coverage through the space. In addition to the information required for Algorithm \ref{alg:DMPP}, we also need our starting position which is assumed to be on or inside the polygon. We take the cell corners produced in the DMPP algorithm and move these away from the polygon edges towards the inside of the polygon by $\delta$ at a direction orthogonal to the sweep direction $\psi_{sd}$. This gives us corners, $shrunkCellCnrs$ which are spaced at the desired distance away from the boundary we have already traced in Algorithm \ref{alg:FFCB}. 

We now calculate the transit path from our current position $pos$ to the closest corner of $shrunkCellCnrs$. This is done in 2 steps. Firstly we calculate the euclidean distance between all corners and the current position. We then find the shortest path, and check whether it is inside the polygon. If so, then we have a best solution and we use the primitive procedure genWayPoints() to generate a series of way points spaced $\delta$ apart between the start and end points and set this as the transit path. This allows us to quickly find transit paths between adjoining cells before resorting to more computationally intensive searches. If it is not, we conduct a series of A* searches \cite{Dechter1985}, using the euclidean distance as a heuristic, from $pos$ to all the corners in $shrunkCellCnrs$. We find the shortest length path and set it as the transit path and append it to the lawnmower path. The cell which we transit to is popped from the list $shrunkCellCnrs$, the corresponding cell polygon $cell$ is popped from $cells$, the entry corner index is recorded as $entryCnr$, and we set $pos$ to the position of this entry corner.

Within this current cell we now create a lawnmower path by ray tracing lines orthogonal to the sweep direction, at a spacing of $\delta$, starting at the entry corner position. We shrink these resultant line segments by $\delta$ from the cell boundary, create waypoints on them $\delta$ apart, and join them together. 

\begin{figure}[h]
\centering
\includegraphics[scale = 0.5, natwidth=658 , natheight=532, trim=0 0 0cm 0]{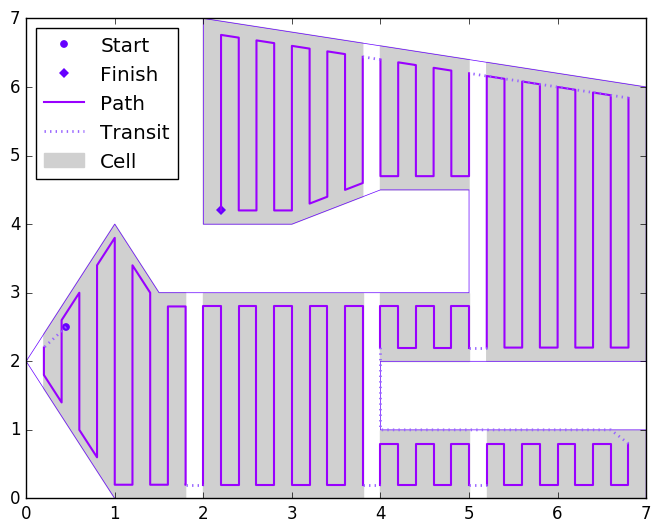}
\caption{DMPP and Path Generation}
\label{fig:DMPPeg}
\end{figure}

Once we have created the lawnmower path for the current cell we return to the transit path loop to find the path to the next nearest cell and continue until all cells are covered. At which point we return the final path for coverage of the entire polygon. For even the simple example shown in Figure \ref{fig:boustComp} the BCD method results in total within cell path 6\% longer and transit paths 106\% longer for a total path which is 12\% longer than our DMPP and path generation algorithm. An example for a more complex polygon can be seen in Figure \ref{fig:DMPPeg}. Unlike the BCD in which the union of the cells is the polygon, in the DMPP there are spaces between the cells. However these are designed such that the tracklines are exactly upon these edges and thus we achieve the even coverage we desire. 

Whilst this is a greedy method in that it is only ever looking ahead one cell, it is superior to the  BCD, which naively takes the nearest anti-clockwise cell regardless of transit distance. The resultant path length could be improved at the expense of computation time by a deeper search, though an exhaustive search would quickly become infeasible as the number of cells grows due to the NP nature of of the problem. The sweep direction could also be optimised, perhaps through running a number of potential rotations similar to the trapezoidal sweep optimisation in \cite{Oksanen2009}, though unlike their method, we must choose the same trackline orientation for all cells as they have been created monotone to the same sweepline orientation. Whilst we have only shown the DMPP and path fitting algorithms for polygons without loops or holes, as this was our use case, the algorithms themselves can easily be adapted to this, which would require some of the primitive polygon procedures such as traceBoundary() to be adjusted.

\section{Simulation}

A simulated Bathymetry has been created to demonstrate these algorithms. A vessel with perfect localisation, sensing and control is tested to focus on validating the performance of the coverage algorithm itself. The parameter settings can be seen in Table \ref{tab:Params}. From the start point the vehicle is driven in a circle of radius 5m for 50s to gain some initialisation points for the GP. After this initialisation, the GP hyper-parameters are estimated, and then again every 30s. 

\begin{table}[h]
\centering
	\begin{tabular}{l|c|c}
	& Simulation & Field Trial \\ \hline
	Velocity (m/s) & 1.0 & $\approx$ 1.0 \\
	$z_t$ (m) & 4.5 & 4.0 \\
	$r$ (m) & 2.5, 5.0, 7.5 & 5.0 \\
	$\delta$ (m) & 10.0 & 5.0 \\
	$\psi_{sd}$ (rad) & 0.0 & 0.0 \\
	Start Point & 250E, 350N & 0E, 0N \\
	Hyper-Parameter & & \\
	re-estimation time (s) & 30 & 30 \\
	IMU (Hz) & 1 & 50 \\
	GPS (Hz) & 1 & 1 \\
	Control Loop (Hz) & 1Hz & 5Hz \\
	$\lambda_{EMA}$ (s) & 5 & 5 \\	
	\end{tabular}
\caption{Parameter Settings for Simulation and Field Trials}
\label{tab:Params}	
\end{table}

\begin{figure}[h]
	\centering
	\includegraphics[scale = 0.55, natwidth=1255 , natheight=563,trim=1cm 0 0cm 0cm]	{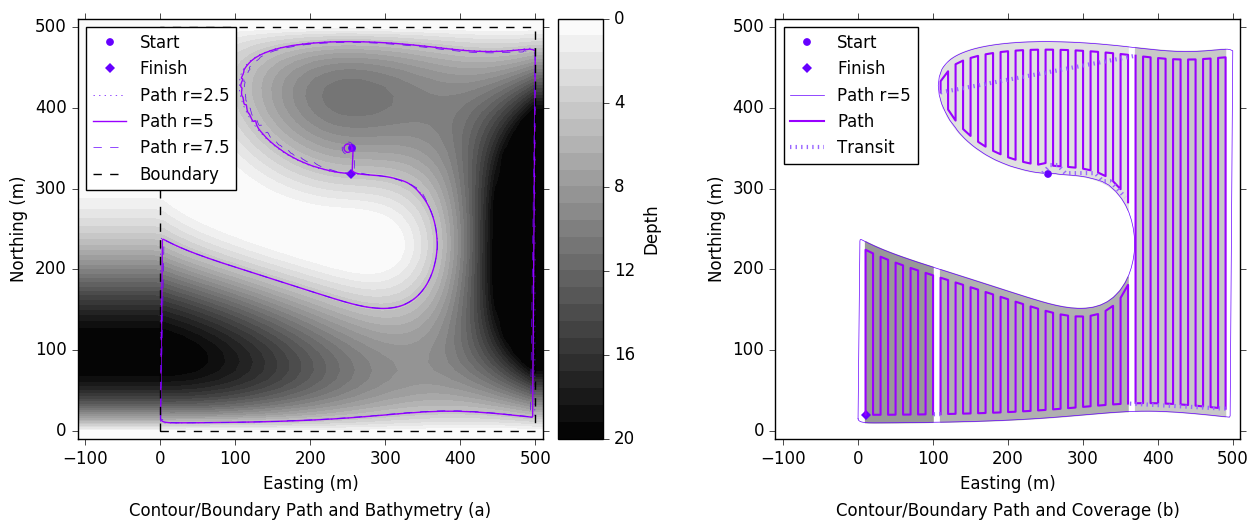}
	\caption{Simulation}
	\label{fig:sim}
\end{figure}

As can be seen in Figure \ref{fig:sim}(a) the vessel follows the contour gradient it has discovered by searching on its GP model of the bathymetry and follows south until it arrives at the target depth. It then turns east and follows this contour until it gets to the western boundary at [0,237]. At this point it follows the boundary south into deeper water until this boundary following would take it shallower than the target depth at which point it turns east again and follows the contour. After another boundary and contour following section it completes tracing the intersection of the boundary and the depth contour. 

The algorithm now moves on to partitioning this intersection and creating a path for coverage. Figure \ref{fig:sim} (b) shows the result. The intersection is split into 4 cells. From the start point the closest cell corner, on the south west, cannot be transited to in a straight line as this would take us out of the polygon. An A* Path is generated to take us to this point. A lawnmower path is then generated to take us through this cell to the west. A transit path from the end of this cell is then generated to take us to the north west corner of the next cell, which can be done directly as this stays within the polygon. The lawn mower path is then generated for this cell and the the process repeated for the final 2 cells until we have a plan for coverage of the entire space. 

The main design parameters for these algorithms are the target depth $z_t$, the path spacing $\delta$ and the search radius $r$. The target depth should be chosen based on a combination of the safe operating depth of the vehicle and the depths of interest for the study. The path spacing should be chosen based on the coverage density desired. The search radius impacts the operation of the algorithm in a number of ways. A larger search radius expands the search horizon, though if this is too large it may move us away from where we have certainty in our model. As this search radius increases the ability of the vessel to smoothly follow tight turns in the contour is reduced, and a tracking error between what is directly under the vehicle compared to the depth at the planning horizon whilst following a curve is introduced. As such this parameter should be bounded from above based on an expectation of the minimum radius of curves in the contours it is following. On the lower range of this variable we want the planning horizon to be longer than the distance covered by the vessel between planning points (in the simulation case this is 1m due to a velocity of 1m/s and a control loop of 1Hz). Empirical testing has shown the solution to be robust to the choice of $r$ as can be seen in Figure \ref{fig:sim} (a) where setting $r$ at 2.5m or 7.5m results in a very similar path to $r$ = 5m. 

\section{Field Tests}

\begin{figure}[h]
	\centering
	\includegraphics[scale = .6, natwidth=1085 , natheight=727,trim=0 0 0 0]	{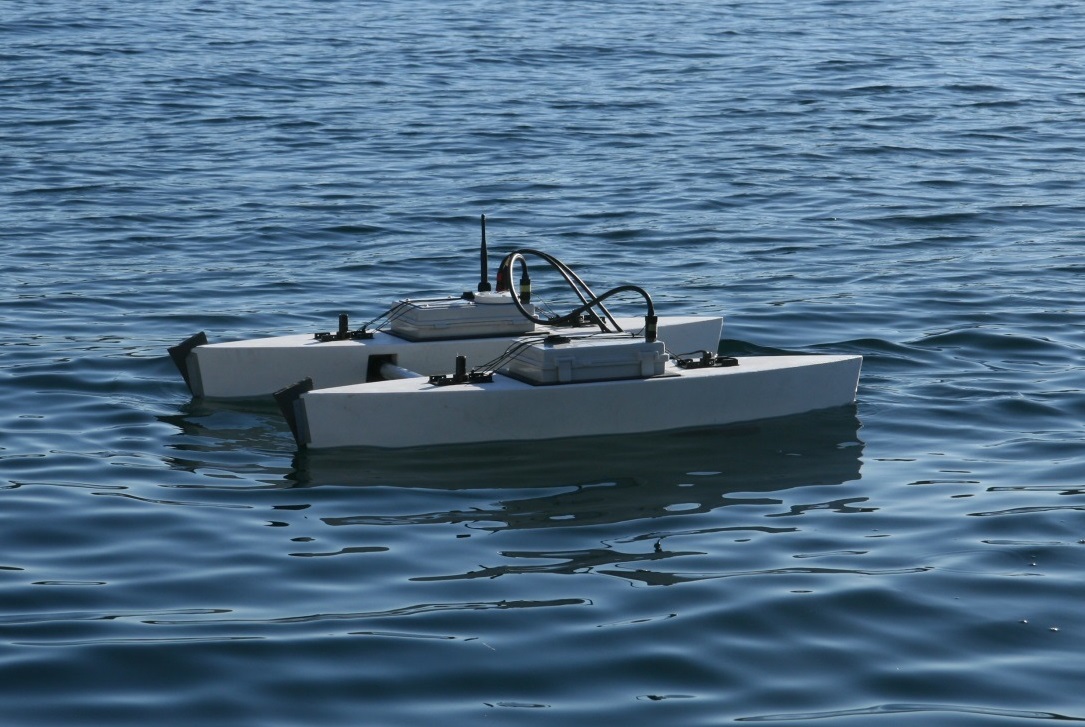}
	\caption{Autononomous Surface Vessel}
	\label{fig:ASV}
\end{figure}

To demonstrate the robustness of these algorithms to the noise introduced from uncertainty in sensing and localisation from operating in the field and the computational limits imposed by an embedded CPU and real time operation, a small autonomous surface vessel was built. The vessel, which can be seen in Figure \ref{fig:ASV}, is a twin hull, differential thrust design. It has a footprint of 1.2m x 0.85m and weighs approximately 10kg. This small form size enables deployment by a single person. Localistation is provided by a Flexpac G6 GPS running at 1Hz combined with a Vectornav VN-100 rugged IMU running at 50Hz fused with a Kalman Filter. Bathymetric sensing  is provided by an Airmar Single Beam Sonar operating at 1Hz. An Embedded Solutions ADLN2000PC  containing an Intel Atom N2600 1.6GHz dual (4 virtual) core processor with 2Gb RAM provides the computational resources. Information relay and manual over-ride control are handled with an Xbee Pro 2.4Ghz RF module. Power is supplied by an Ocean Server battery controller connected to two 6.6Ah Li-Ion battery packs providing 6hrs of operational time. There are also leak and temperature sensors inside the payload boxes, which combined with a heart beat published over the RF channel trigger automatic kill switches for safety. The vessel has a maximum velocity of approximately 1m/s. 

\begin{figure}[t]
	\centering
	\includegraphics[scale = 0.37, natwidth=1544 , natheight=612,trim=0 0 0cm 0cm]	{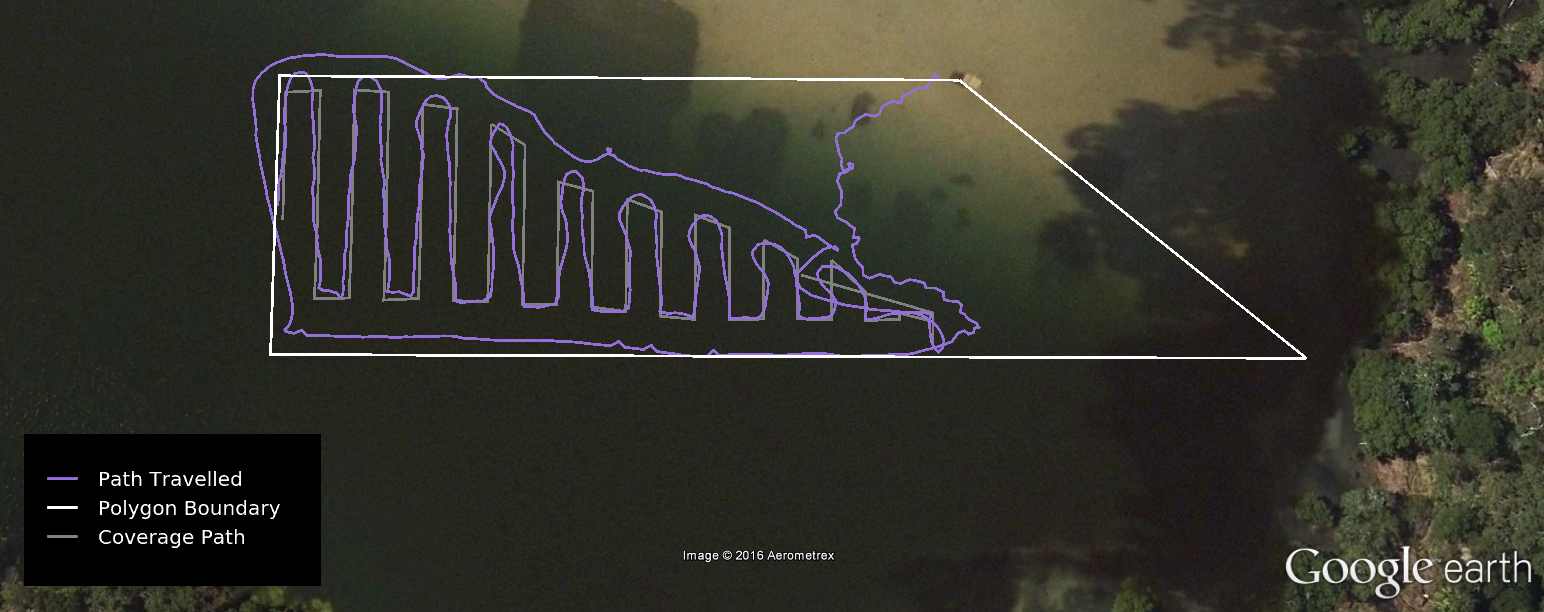}
	\caption{Field Test - Autonomous Bathymetric Sensing Within Depth and Boundary Constraints}
	\label{fig:satImage}
\end{figure}

The framework used for the code running on the ASV is Robot Operating System (ROS) Indigo \cite{quigley2009ros}, installed on Ubuntu(14.04). All code is written in Python for quick development, however ROS also supports C++, so modules can be re-written if processing speed becomes an issue. ROS inherently supports multi-threaded operation with a single launch file initiating the various modules referred to as nodes. Communication between the nodes is handled by both a parameter server to read/write global variables and a publish/subscribe messaging system.

The area chosen for this task is a secluded part of the Port Hacking river, to the south of Sydney, Australia, called Cabbage Tree Basin. This in an interesting area for a number of reasons. The Port Hacking river was the first estuary in Australia closed to commercial fishing in the late 19th century. It is bordered on one side by a residential area, and on the other by the Royal National Park, which is the second oldest national park in the world (after Yellowstone), established in 1879. Cabbage tree basin itself is a significant area of heritage value with a long history of aboriginal occupation. It was the site of the first marine hatchery in Australia in 1900 and is one of the earliest described estuarine wetland areas in Australia \cite{west2000cabbage}. The long shallow entrance to the basin also significantly limits access to recreational boating traffic, enabling unobstructed operation of the ASV.

\begin{figure}[h!]
\subfloat[]{
	\centering
	\includegraphics[natwidth=1348 , natheight=833, width = 0.5\textwidth, trim=0 0 0 0]{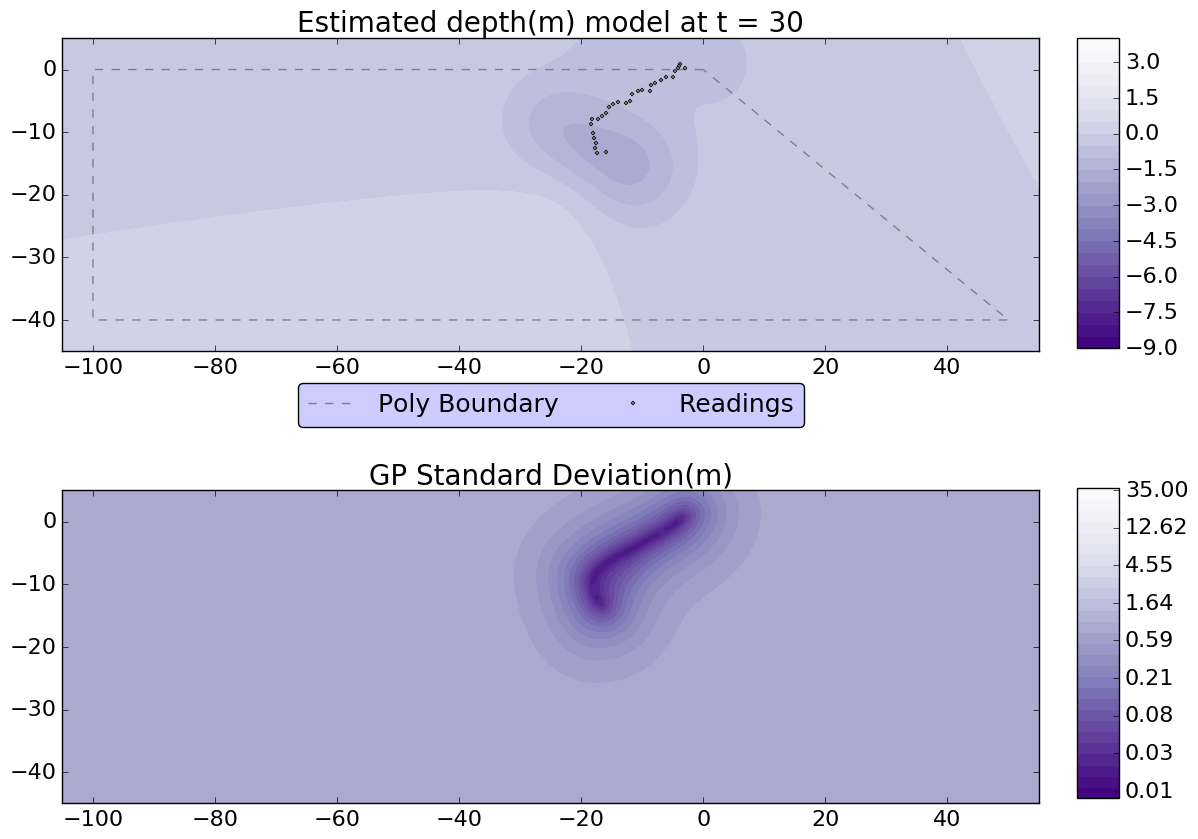}} \label{fig:field_a}
\subfloat[]{
	\centering
	\includegraphics[natwidth=1348 , natheight=833, width = 0.5\textwidth, trim=0 0 0 0]{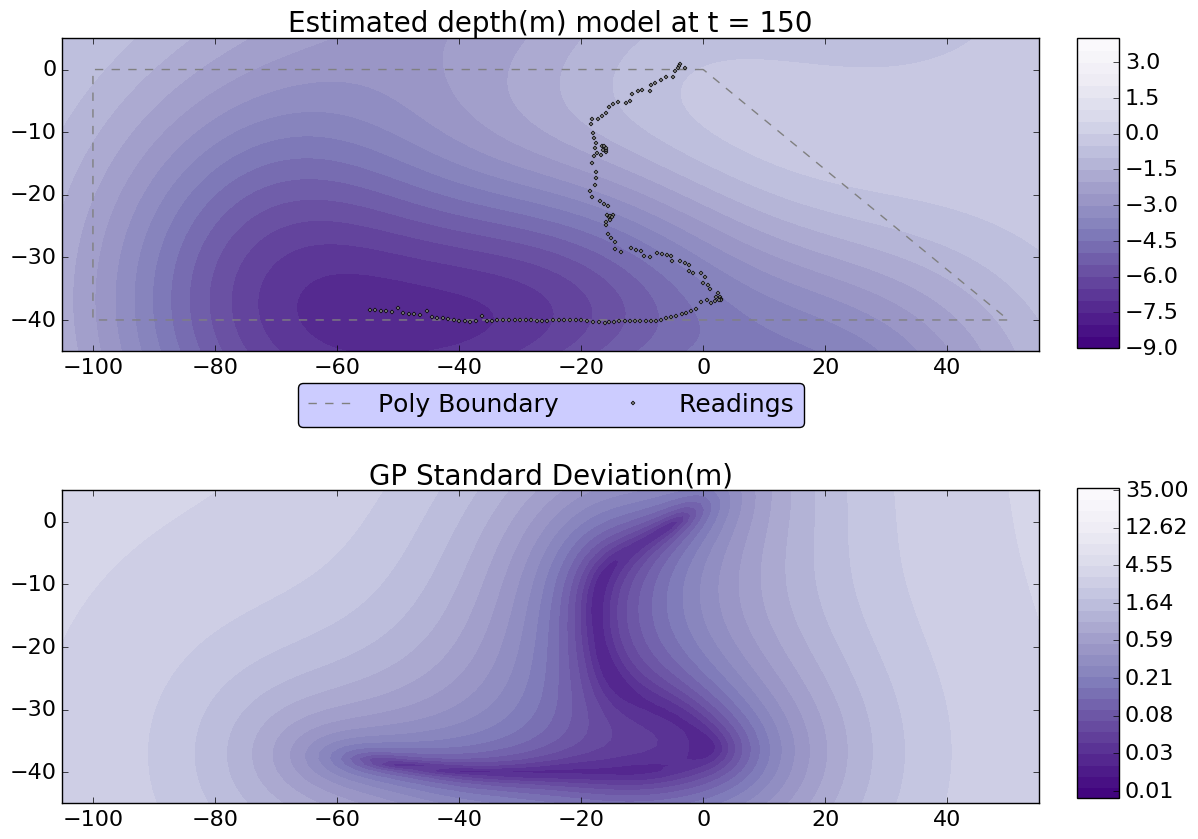}} \label{fig:field_b}
	\\
\subfloat[]{
	\centering
	\includegraphics[natwidth=1348 , natheight=833, width = 0.5\textwidth, trim=0 0 0 0]{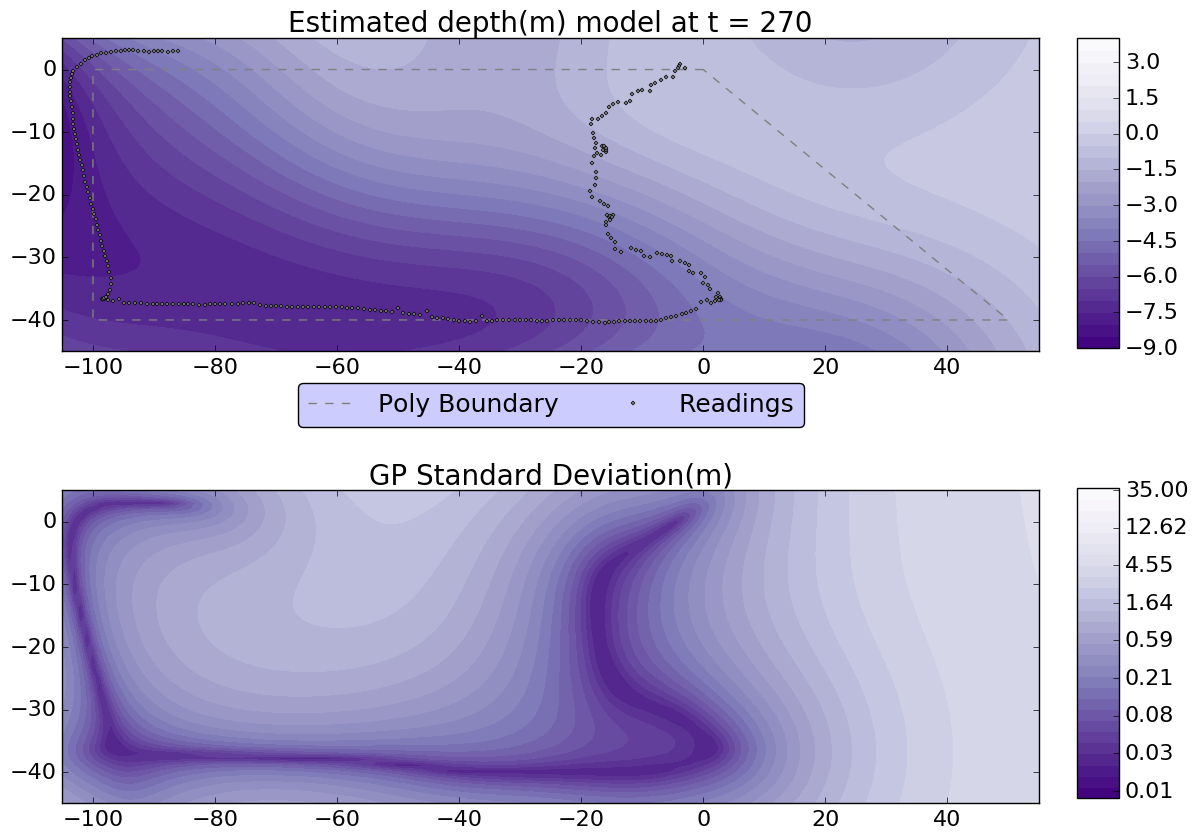}} \label{fig:field_c}
\subfloat[]{
	\centering 
	\includegraphics[natwidth=1348 , natheight=833, width = 0.5\textwidth, trim=0 0 0 0]{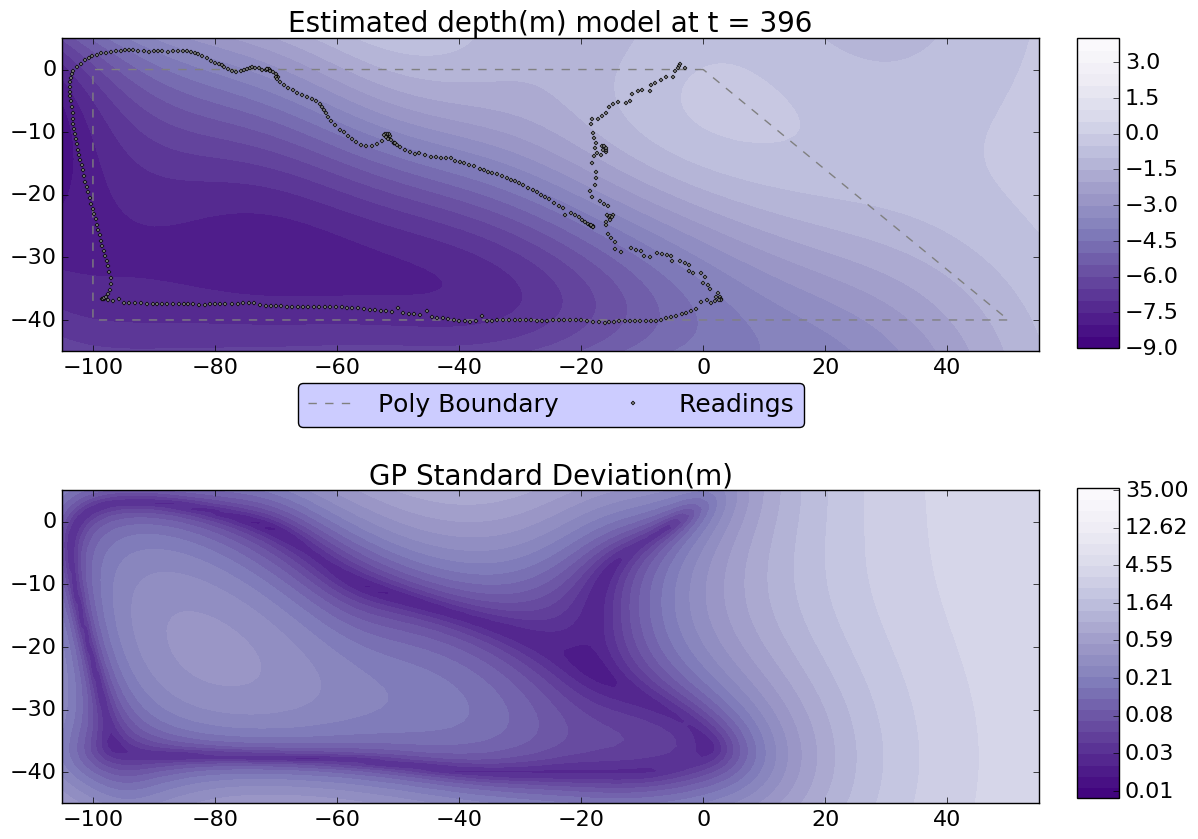}} \label{fig:field_d}
	\\
\subfloat[]{
	\centering
	\includegraphics[natwidth=1348 , natheight=833, width = 0.5\textwidth, trim=0 0 0 0]{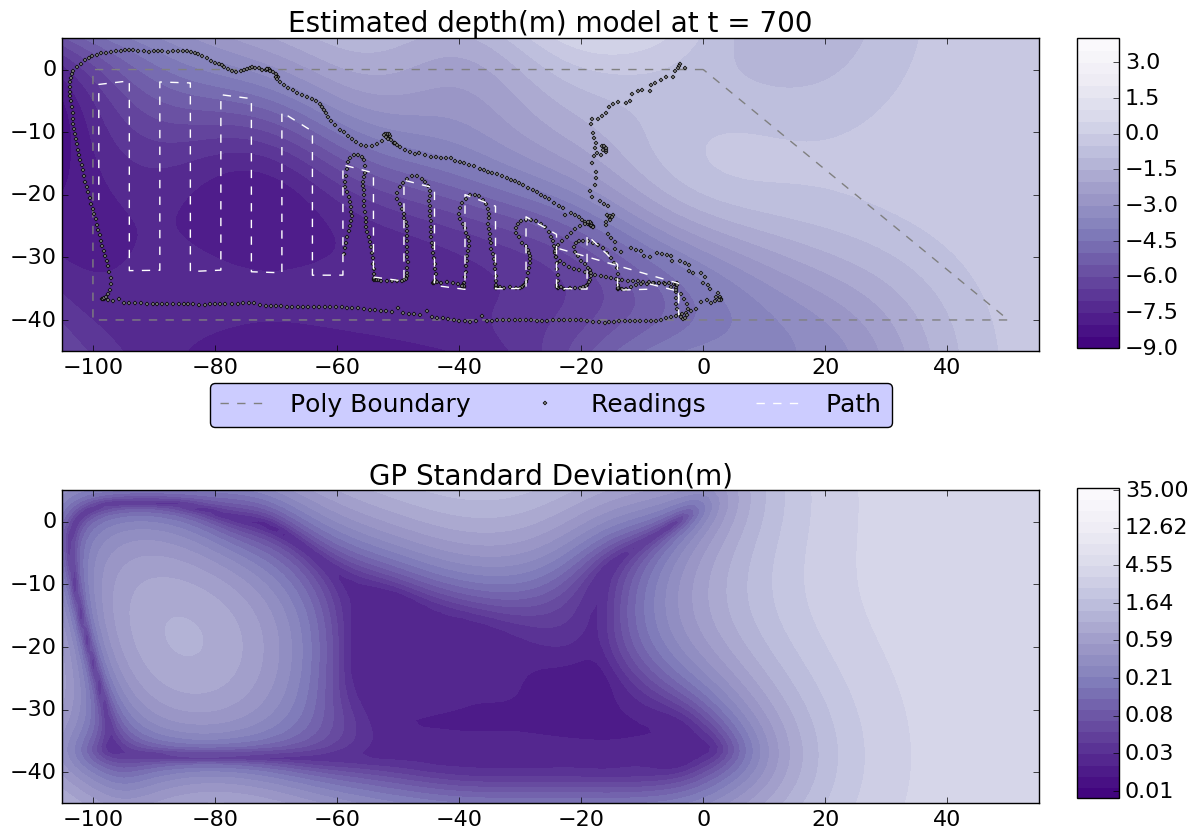}} \label{fig:field_e}
\subfloat[]{
	\centering
	\includegraphics[natwidth=1348 , natheight=833, width = 0.5\textwidth, trim=0 0 0 0]{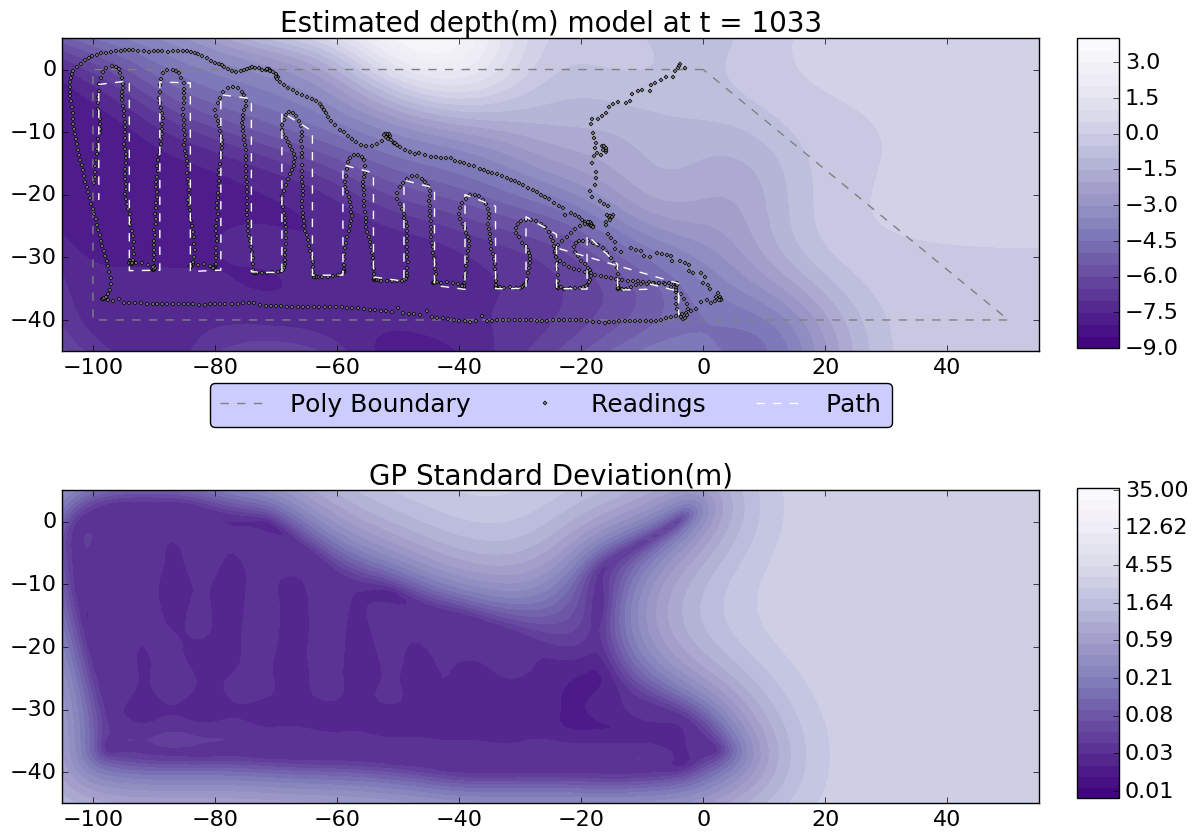}} \label{fig:field_f}
\caption{Path Overlaid on GP Estimated Depth Contours and Variances}
\label{fig:field}
\end{figure}

A bounding polygon 100-150m wide by 40m high was set, as can be seen by the white trapezoid in figure \ref{fig:satImage}. This area encompassed depths from less than 50cm to 8m. The parameters used can be seen in Table \ref{tab:Params}. The mission starts at the North-East corner of the white trapezoid which can be seen in Figure \ref{fig:satImage}. The vessel was manually driven south-west for approximately 10m until it reached a depth of 1m. It was then driven in an arc for 5s (achieving a quarter of a circle). The GP hyper-parameters are estimated from these initial points, and then again every 30s. The FFCB control loop, Algorithm \ref{alg:FFCB}, then started operation. Figure \ref{fig:field} shows some snapshots of the GP model, planning and path travelled during the mision. In each of the 6 pairs of figures the upper figure shows the path of the vessel overlaid, at a given point in time, on the bathymetry estimated by the model, whilst the lower chart shows the confidence of the model at this time point through the standard deviation (in log scale), with 1 data point per second. A video showing the evolution of the model as each new data point arrives can also be seen online \cite{Wilson2015gpvideo}.

To increase the robustness of the algorithm in the field trials we used an exponentially smoothed average of the current estimated heading, with a half life of 5s (equivalent to the time taken to cross the planning horizon). This was used in contour following mode to center the search space for the roseSolve() algorithm to smooth out the short term effects of any environmental forcing on the instantaneous heading.

Figure \ref{fig:field}(a) shows the path and model immediately after the initialisation period and first hyper-parameter estimation. At this point the algorithm does not have a very good model to work with when trying to follow the contour. Due to this the vessel traces a tight circle between t=30s an t=37s before continuing to follow the contour south until it hits the desired depth. It then follows this depth contour south east until it gets to the southern boundary. This boundary is followed along the bottom edge, up the western side and partially across the top until this would take it too shallow, as can be seen in Figure \ref{fig:field}(c), at which point it switches back to contour following mode with the vessel then following the contour back toward where it first found the contour. This is achieved by t=396s. The intersection of the depth countour and boundary polygon is now used by the DMPP and path generation algorithm to plan a path. In this case the intersection is already a monotone polygon with respect to the sweep direction and thus only 1 cell was created and a lawnmower path was fit to it. The vessel then followed this path for coverage. As can be seen from the Figures \ref{fig:field}(d) to \ref{fig:field}(f), the standard deviation of the map within the intersection falls to around 3cm after the coverage task has been completed at point 1033.

\begin{figure}[h]
	\centering
	\includegraphics[scale = 0.6, natwidth=768 , natheight=380,trim=0 0 0 0]	{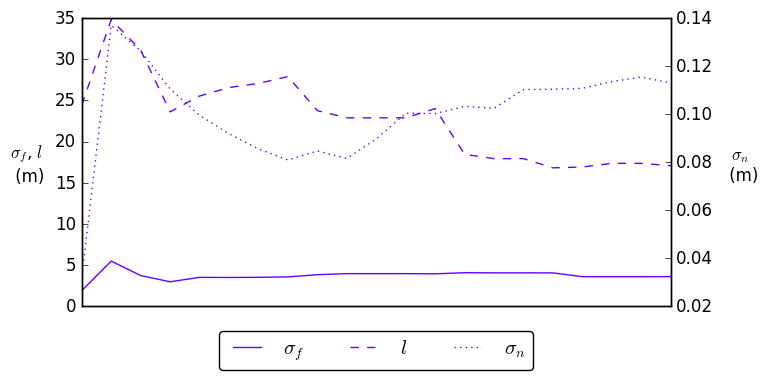}
	\caption{Iterative Hyper-Parameter Estimation}
	\label{fig:HPs}
\end{figure}

The history of the hyperparameters as they were optimised throughout the mission can be seen in Figure \ref{fig:HPs}. The range of the noise standard deviation, whilst it looks large, is on a different scale and 2 orders of magnitude smaller than the model noise, so is irrelevant in terms of its implications for the control algorithm.  The model standard deviation, $\sigma_f$, relatively quickly stabalises around 3m to 4m by the 3rd point. The characteristic length scale estimate slowly decreases as we get more data, though from the 4th point the estimate is in a range of 16.8 to 27.9m. The final estimated hyper-parameters are $\sigma_f = 3.6$m, $l = 17.07$m and $\sigma_n = 0.115$m. In spite of this variability in the hyper-parameter estimates throughout the mission, the algorithm showed its ability to follow an estimated contour, which appears reasonable when compared to the final estimate of the bathymetry as can be seen in Figure \ref{fig:field}(f).

On the day of the test there was a moderate breeze from the south east. This resulted in the vessel straying slightly from the boundary path whilst in boundary following mode as can be seen in Figure \ref{fig:satImage} where the path traveled is slightly inside the polygon boundary on the south west corner and slightly outside on the north west corner. Superior tracking performance in the presence of external forces such as winds and currents could likely be achieved with a more sophisticated controller than the proportional controller used on the ASV. Another problem occasionally encountered in control of the vessel was the presence of floating seaweed becoming tangled in the propellers, in the worst cases requiring a restart of the mission. Whilst the propellers do have a cage around them for safety, a finer mesh could perhaps help with deflecting the seaweed before it becomes entangled around the prop shaft.

The choice of a relatively short range greedy search to decide on the path for the vessel to take is seen to be justified by the relatively local nature of the information available. Given we are only sensing directly downwards, there is a limit to how far we can confidently project this information. This can be seen by comparing the sequential plots in Figure \ref{fig:field}. In Figure \ref{fig:field}(b) the model believes that the depth starts to rise about 7m away from the current position, whereas once this area has been sampled, which we can see in Figure \ref{fig:field}(c), we can see that it actually stayed constant. If we had tried to make a plan to follow the whole contour from the data and model we had at the time of Figure \ref{fig:field}(b), we would not have achieved the result seen in Figure \ref{fig:field}(c) by following the contour in a local region.  

\section{Conclusions and Future Work}
In this paper we have developed and implemented a suite of algorithms for autonomously finding and following the intersection of a bathymetric contour and a bounding polygon, and then fitting a path for coverage within this boundary for the purpose of producing a map of the bathymetry in an unknown area. We provide a new algorithm for the partitioning of complex polygonal workspaces and the planning of coverage paths within them, which is more efficient than the BCD method, and explicitly handles the complex boundary shapes. We have implemented computationally efficient methods for the updating of cholseky matrices used by the GP to allow online fitting and prediction of a bathymetric map, including online optimisation of the hyper-parameters of the GP. Code for all algorithms and the implementation on the ASV are provided on GitHub \cite{WilsonASVGit}.

We have also developed and implemented a small autonomous surface vehicle with which we can test this and other robotics algorithms and conduct surveys in estuarine water ways. We tested these algorithms in simulation and in the field and through these experiments have shown the robustness of the platform and algorithms to uncertainty introduced by sensor noise and environmental forcing in a dynamic environment combined with the ability to run in real time on a small embedded system.

GP's inherently have scaling issues for fitting and prediction due to the inversions required of the covariance matrices which are $O(n^3)$. This was not an issue for our surveys, but would need to be considered for significant increases in scale. Future work in this area could look at implementing methods for sparsification or reduction of the covariance matrices \cite{hensman2012gaussian}. Specific to our algorithms, given the search space is relatively local, we could use only the data points relevant for making a map in the area searched. This could be guided by the model such that the area within which we should use the data points would be guided by the combination of the search range and the hyper-parameters which determine the range of predictive power of each data point. Computational speed for estimation of the hyper-parameter is less of an impediment for the online autonomous operation as it can be run independently of the control loop in a separate thread, or potentially even off board the ASV if the data points are all transmitted across, with the results integrated as they arrive.

In this study we  mapped out a simple deterministic path inside the mapped boundary. A natural extension would be to use the variance information from the GP to drive the exploration within this boundary to create maps of a desired fidelity, with a minimum length path. A non-stationary covariance matrix could be fit which could allow heterogeneous spacing of sampling points dependent on the local variability of the bathymetry.

The single beam sonar, whilst likely to be replaced by a multi-beam sonar for creating maps for navigational purposes, serves as a useful analogue for other variables of interest in estuarine environments which can only be measured with point sensors. We intend in future work to use point sensors for Dissolved Oxygen, Salinity, Water Temperature, Tubidity, Ph and Chlorophyll-a, to measure the health of estuaries. Spatio-temporal GP's could autonomously model these variables in an estuary over the cycle of an incoming tidal front. The path of the ASV will be planned online as the model is fit to create the best model of these variables.

\subsubsection*{Acknowledgments}
This work is supported by the Department of Education and Training through the Australian Postgraduate Awards (APA) scheme and the Australian Research Council (ARC) through the Future Fellowships (FT) award. The authors would also like to thank Seatools Pty Ltd for support in construction of the platform used for the field trials.

\bibliographystyle{apalike}

\begin{thebibliography}{}

\end{thebibliography}


\begin{thebibliography}{}

\bibitem[Acar and Choset, 2001]{Acar2001}
Acar, E.~U. and Choset, H. (2001).
\newblock Robust sensor-based coverage of unstructured environments.
\newblock In {\em 2001 IEEE/RSJ International Conference on Intelligent Robots
  and Systems}, volume~1, pages 61--68. IEEE.

\bibitem[Acar et~al., 2002]{Acar2002}
Acar, E.~U., Choset, H., Rizzi, A.~A., Atkar, P.~N., and Hull, D. (2002).
\newblock {Morse decompositions for coverage tasks}.
\newblock {\em The International Journal of Robotics Research}, 21(4):331--344.

\bibitem[Arkin et~al., 2000]{Arkin2000}
Arkin, E.~M., Held, M., and Smith, C.~L. (2000).
\newblock {Optimization Problems Related to Zigzag Pocket Machining}.
\newblock {\em Algorithmica}, 26:197--236.

\bibitem[Bender et~al., 2010]{Bender2010}
Bender, A., Williams, S.~B., Pizarro, O., and Jakuba, M.~V. (2010).
\newblock Adaptive exploration of benthic habitats using gaussian processes.
\newblock In {\em OCEANS 2010}, pages 1--10. IEEE.

\bibitem[Choset and Pignon, 1998]{Choset1998}
Choset, H. and Pignon, P. (1998).
\newblock {Coverage Path Planning : The Boustrophedon Cellular Decomposition}.
\newblock {\em Field and Service Robotics}, pages 203--209.

\bibitem[Clark et~al., 2013]{Clark2013a}
Clark, C.~M., Hancke, K., Xydes, A., Hall, K., Schreiber, F., Klemme, J., Lehr,
  J., and Moline, M. (2013).
\newblock {Estimation of volumetric oxygen concentration in a marine
  environment with an autonomous underwater vehicle}.
\newblock {\em Journal of Field Robotics}, 30(1):1--16.

\bibitem[Dechter and Pearl, 1985]{Dechter1985}
Dechter, R. and Pearl, J. (1985).
\newblock Generalized best-first search strategies and the optimality of a*.
\newblock {\em J. ACM}, 32(3):505--536.

\bibitem[Dunbabin et~al., 2009]{Dunbabin2009}
Dunbabin, M., Grinham, A., and Udy, J. (2009).
\newblock An autonomous surface vehicle for water quality monitoring.
\newblock In {\em Australasian Conference on Robotics and Automation (ACRA)},
  pages 2--4.

\bibitem[Ferreira et~al., 2007]{Ferreira2007}
Ferreira, H., Martins, R., Marques, E., Pinto, J., Martins, a., Almeida, J.,
  Sousa, J., and Silva, E.~P. (2007).
\newblock {SWORDFISH: an Autonomous Surface Vehicle for Network Centric
  Operations}.
\newblock {\em OCEANS 2007 - Europe}, pages 1--6.

\bibitem[Ferri et~al., 2008]{Ferri2008}
Ferri, G., Jakuba, M.~V., and Yoerger, D.~R. (2008).
\newblock {A novel method for hydrothermal vents prospecting using an
  autonomous underwater robot}.
\newblock In {\em 2008 IEEE International Conference on Robotics and
  Automation}, pages 1055--1060. Ieee.

\bibitem[Galceran and Carreras, 2013]{Galceran2013}
Galceran, E. and Carreras, M. (2013).
\newblock {A survey on coverage path planning for robotics}.
\newblock {\em Robotics and Autonomous Systems}, 61(12):1258--1276.

\bibitem[Grasmueck et~al., 2006]{Grasmueck2006}
Grasmueck, M., Eberli, G.~P., Viggiano, D.~a., Correa, T., Rathwell, G., and
  Luo, J. (2006).
\newblock {Autonomous underwater vehicle (AUV) mapping reveals coral mound
  distribution, morphology, and oceanography in deep water of the Straits of
  Florida}.
\newblock {\em Geophysical Research Letters}, 33(23):1--6.

\bibitem[Hensman and Lawrence, 2012]{hensman2012gaussian}
Hensman, J. and Lawrence, N. (2012).
\newblock Gaussian processes for big data through stochastic variational
  inference.
\newblock {\em Adv Neural Inf Process Syst}, 25.

\bibitem[Hine et~al., 2009]{hine2009wave}
Hine, R., Willcox, S., Hine, G., and Richardson, T. (2009).
\newblock The wave glider: A wave-powered autonomous marine vehicle.
\newblock In {\em OCEANS 2009}. IEEE.

\bibitem[Hitz et~al., 2012]{hitz2012}
Hitz, G., Pomerleau, F., Garneau, M.-E., Pradalier, C., Posch, T., Pernthaler,
  J., and Siegwart, R.~Y. (2012).
\newblock Autonomous inland water monitoring: Design and application of a
  surface vessel.
\newblock {\em Robotics \& Automation Magazine, IEEE}, 19(1):62--72.

\bibitem[Hollinger et~al., 2012]{Hollinger2012}
Hollinger, G.~a., Englot, B., Hover, F.~S., Mitra, U., and Sukhatme, G.~S.
  (2012).
\newblock {Active planning for underwater inspection and the benefit of
  adaptivity}.
\newblock {\em The International Journal of Robotics Research}, 32(1):3--18.

\bibitem[Johnson-Roberson et~al., 2010]{JohnsonRobertson2010}
Johnson-Roberson, M., Pizarro, O., Williams, S.~B., and Mahon, I. (2010).
\newblock {Generation and Visualization of Large-Scale Three-Dimensional
  Reconstructions from Underwater Robotic Surveys}.
\newblock {\em Journal of Field Robotics}, 27(1):21--51.

\bibitem[Kamon and Rivlin, 1997]{Kamon1997}
Kamon, I. and Rivlin, E. (1997).
\newblock {Sensory-based motion planning with global proofs}.
\newblock {\em IEEE Transactions on Robotics and Automation}, 13(6):814--822.

\bibitem[Krause, 2008]{Krause2008}
Krause, A. (2008).
\newblock {Near-Optimal Sensor Placements in Gaussian Processes : Theory ,
  Efficient Algorithms and Empirical Studies}.
\newblock {\em The Journal of Machine Learning Research}, 9:235--284.

\bibitem[Latombe, 2012]{latombe2012robot}
Latombe, J.-C. (2012).
\newblock {\em Robot motion planning}, volume 124.
\newblock Springer Science \& Business Media.

\bibitem[Lawrance and Sukkarieh, 2011]{Lawrance2011}
Lawrance, N.~R. and Sukkarieh, S. (2011).
\newblock {Path planning for autonomous soaring flight in dynamic wind fields}.
\newblock {\em 2011 IEEE International Conference on Robotics and Automation},
  pages 2499--2505.

\bibitem[Manley et~al., 2000]{Manley2000}
Manley, J.~E., Marsh, A., Comforth, W., and Wiseman, C. (2000).
\newblock {Evolution of the Autonomous Surface Craft AutoCat}.
\newblock In {\em OCEANS 2000}, pages 403--408. IEEE.

\bibitem[Marchant and Ramos, 2012]{Marchant2012}
Marchant, R. and Ramos, F. (2012).
\newblock {Bayesian optimisation for Intelligent Environmental Monitoring}.
\newblock {\em 2012 IEEE/RSJ International Conference on Intelligent Robots and
  Systems}, pages 2242--2249.

\bibitem[Marchant et~al., 2014]{Marchant2014}
Marchant, R., Ramos, F., and Sanner, S. (2014).
\newblock Sequential bayesian optimisation for spatial-temporal monitoring.
\newblock In {\em Proceedings of the Thirtieth Conference Annual Conference on
  Uncertainty in Artificial Intelligence (UAI-14)}, pages 553--562, Corvallis,
  Oregon. AUAI Press.

\bibitem[Matos and Cruz, 2008]{Matos2008}
Matos, A. and Cruz, N. (2008).
\newblock {Positioning control of an underactuated surface vessel}.
\newblock {\em OCEANS 2008}, pages 1--5.

\bibitem[Naeem et~al., 2012]{Naeem2012}
Naeem, W., Sutton, R., and Xu, T. (2012).
\newblock {An integrated multi-sensor data fusion algorithm and autopilot
  implementation in an uninhabited surface craft}.
\newblock {\em Ocean Engineering}, 39:43--52.

\bibitem[Nguyen et~al., 2013]{Nguyen2013}
Nguyen, J., Lawrance, N., Fitch, R., and Sukkarieh, S. (2013).
\newblock {Energy-Constrained Motion Planning for Information Gathering with
  Autonomous Aerial Soaring}.
\newblock In {\em 2013 IEEE International Conference on Robotics and
  Automation}, pages 3810--3816. IEEE.

\bibitem[Oksanen and Visala, 2009]{Oksanen2009}
Oksanen, T. and Visala, A. (2009).
\newblock {Algorithms for Agricultural Field Machines}.
\newblock {\em Journal of Field Robotics}, 26(8):651--668.

\bibitem[Osborne et~al., 2008]{Osborne2008}
Osborne, M.~a., Rogers, A., Ramchurn, S., Roberts, S.~J., and Jennings, N.~R.
  (2008).
\newblock {Towards Real-Time Information Processing of Sensor Network Data
  using Computationally Efficient Multi-output Gaussian Processes}.
\newblock {\em International Conference on Information Processing in Sensor
  Networks (IPSN 2008)}, pages 109--120.

\bibitem[Pascoal et~al., 2000]{Pascoal2000}
Pascoal, A., Oliveira, P., Silvestre, C., Sebasti{\~{a}}o, L., Rufino, M.,
  Barroso, V., Gomes, J., Seube, N., Champeau, J., Dhaussy, P., Sauce, V., and
  Moiti{\'{e}}, R. (2000).
\newblock {Robotic Ocean Vehicles for Marine Science Applications : the
  European ASIMOV Project}.
\newblock In {\em OCEANS 2000}. IEEE.

\bibitem[Quigley et~al., 2009]{quigley2009ros}
Quigley, M., Conley, K., Gerkey, B., Faust, J., Foote, T., Leibs, J., Wheeler,
  R., and Ng, A.~Y. (2009).
\newblock {ROS}: an open-source robot operating system.
\newblock In {\em ICRA workshop on open source software}, volume~3, page~5.

\bibitem[Rasmussen and Williams, 2006]{Williams2006}
Rasmussen, C.~E. and Williams, C. K.~I. (2006).
\newblock {\em {Gaussian Processes for Machine Learning}}.
\newblock The MIT Press.

\bibitem[Saildrone, 2016]{Saildrone}
Saildrone (2016).
\newblock Saildrone.
\newblock \url{www.saildrone.com}.
\newblock Accessed: 2016-01-18.

\bibitem[Senet et~al., 2008]{Senet2008}
Senet, C.~M., Seemann, J., Flampouris, S., and Ziemer, F. (2008).
\newblock {Determination of bathymetric and current maps by the method DiSC
  based on the analysis of nautical X-band radar image sequences of the sea
  surface (November 2007)}.
\newblock {\em IEEE Transactions on Geoscience and Remote Sensing},
  46(8):2267--2279.

\bibitem[Soares et~al., 2013]{Soares2013}
Soares, D., L{\'{u}}cio, C., J{\'{u}}nior, N., and Cunha, W.~C. (2013).
\newblock {Autonomous Navigation of a Small Boat Using IMU / GPS / Digital
  Compass Integration}.
\newblock In {\em Systems Conference (SysCon), 2013 IEEE International}, pages
  468--474.

\bibitem[Souza et~al., 2014]{Souza2014}
Souza, J.~R., Marchant, R., Ott, L., Wolf, D.~F., and Ramos, F. (2014).
\newblock Bayesian optimisation for active perception and smooth navigation.
\newblock In {\em 2014 IEEE International Conference on Robotics and
  Automation}, pages 4081--4087. IEEE.

\bibitem[Tokekar et~al., 2010]{Tokekar2010}
Tokekar, P., Bhadauria, D., Studenski, A., and Isler, V. (2010).
\newblock {A robotic system for monitoring carp in Minnesota lakes}.
\newblock {\em Journal of Field Robotics}, 27(6):779--789.

\bibitem[West and West, 2000]{west2000cabbage}
West, R. and West, J. (2000).
\newblock Cabbage tree basin: Natural values and options for management.
\newblock {\em Unpublished report for the Hacking Catchment Management
  Committee}, 79.

\bibitem[Wilson, 2016]{WilsonASVGit}
Wilson, T. (2016).
\newblock {DMPP, online GP and ASV code}.
\newblock \url{ https://zenodo.org/record/47963}.

\bibitem[Wilson and Williams, 2015]{Wilson2015gpvideo}
Wilson, T. and Williams, S.~B. (2015).
\newblock Adaptive path planning for depth constrained bathymetric mapping.
\newblock \url{https://www.youtube.com/watch?v=yyTfwQdxD9U}.

\bibitem[Yang et~al., 2013]{Yang2013}
Yang, K., {Keat Gan}, S., and Sukkarieh, S. (2013).
\newblock {A Gaussian process-based RRT planner for the exploration of an
  unknown and cluttered environment with a UAV}.
\newblock {\em Advanced Robotics}, 27(6):431--443.

\bibitem[Zhu et~al., 1997]{zhu1997algorithm}
Zhu, C., Byrd, R.~H., Lu, P., and Nocedal, J. (1997).
\newblock Algorithm 778: L-bfgs-b: Fortran subroutines for large-scale
  bound-constrained optimization.
\newblock {\em ACM Transactions on Mathematical Software (TOMS)},
  23(4):550--560.

\end{thebibliography}

\end{document}